\documentclass[11pt]{article}
\usepackage[a4paper, margin=1.25in]{geometry}
\usepackage[T1]{fontenc}
\usepackage{graphicx}
\usepackage[usenames,dvipsnames,svgnames,x11names]{xcolor}
\usepackage{soul}
\newcommand{\hly}[1]{%
    {%
    \sethlcolor{yellow!10}%
    \hl{#1}%
    }%
}
\newcommand{\hlyy}[1]{%
    {%
    \sethlcolor{yellow!40}%
    \hl{#1}%
    }%
}
\newcommand{\hlb}[1]{%
    {%
    \sethlcolor{blue!15}%
    \hl{#1}%
    }%
}
\newcommand{\hlg}[1]{%
    {%
    \sethlcolor{green!20}%
    \hl{#1}%
    }%
}
\newcommand{\hlrr}[1]{%
    {%
    \sethlcolor{red!20}%
    \hl{#1}%
    }%
}

\usepackage[usenames,dvipsnames,svgnames,x11names]{xcolor}

\colorlet{WHITE}{white}
\colorlet{FORESTGREEN}{ForestGreen}
\colorlet{MAROON}{Maroon}

\definecolor{darkblue}{rgb}{0.0,0.0,0.6}
\newcommand{\takeaway}[1]{\noindent{\textcolor{darkblue}{$\blacktriangleright$~{\textsf{#1}}}}}

\usepackage{balance}

\usepackage[T1]{fontenc}
\usepackage[normalem]{ulem}

\newcommand{\para}[1]{{\textbf{#1.~}}}

\usepackage[nocompress]{cite}

\usepackage{listings}

\lstdefinelanguage{XML}
{
basicstyle=\ttfamily\footnotesize,
  morestring=[b]",
  moredelim=[s][\bfseries\color{Maroon}]{<}{\ },
  moredelim=[s][\bfseries\color{Maroon}]{</}{>},
  moredelim=[l][\bfseries\color{Maroon}]{/>},
  moredelim=[l][\bfseries\color{Maroon}]{>},
  morecomment=[s]{<?}{?>},
  morecomment=[s]{<!--}{-->},
  commentstyle=\color{gray},
  stringstyle=\color{blue},
  identifierstyle=\color{red}
}
\definecolor{delim}{RGB}{20,105,176}
\definecolor{numb}{RGB}{106, 109, 32}
\definecolor{string}{rgb}{0.64,0.08,0.08}

\lstdefinelanguage{json}{
    numbers=left,
    numberstyle=\small,
    frame=single,
    rulecolor=\color{black},
    showspaces=false,
    showtabs=false,
    breaklines=true,
    postbreak=\raisebox{0ex}[0ex][0ex]{\ensuremath{\color{gray}\hookrightarrow\space}},
    breakatwhitespace=true,
    basicstyle=\ttfamily\small,
    upquote=true,
    morestring=[b]",
    stringstyle=\color{string},
    literate=
     *{0}{{{\color{numb}0}}}{1}
      {1}{{{\color{numb}1}}}{1}
      {2}{{{\color{numb}2}}}{1}
      {3}{{{\color{numb}3}}}{1}
      {4}{{{\color{numb}4}}}{1}
      {5}{{{\color{numb}5}}}{1}
      {6}{{{\color{numb}6}}}{1}
      {7}{{{\color{numb}7}}}{1}
      {8}{{{\color{numb}8}}}{1}
      {9}{{{\color{numb}9}}}{1}
      {\{}{{{\color{delim}{\{}}}}{1}
      {\}}{{{\color{delim}{\}}}}}{1}
      {[}{{{\color{delim}{[}}}}{1}
      {]}{{{\color{delim}{]}}}}{1},
}

\newcommand{\agen}{AeroGen\xspace}
\newcommand{\CLG}{CLG\xspace}
\newcommand{\adaas}{AeroDaaS\xspace}
\newcommand{\as}{AeroStack2\xspace}

\usepackage{tikz}

\newcommand{\pie}[1]{%
\begin{tikzpicture}
 \draw (0,0) circle (1ex);\fill (1ex,0) arc (0:#1:1ex) -- (0,0) -- cycle;
\end{tikzpicture}%
}

\newcommand{\czero}{\pie{0}}

\newcommand{\cfour}{\pie{360}}

\usepackage{moreverb}
\usepackage[nounderscore]{syntax}

\usepackage[cmex10]{amsmath}
\usepackage{amssymb}
\usepackage{mathtools}
\usepackage{amsfonts}

\usepackage{subfig} 
\usepackage{makecell}

\usepackage{subcaption}
\usepackage{algorithmicx}
\usepackage{algpseudocode}
\usepackage[ruled]{algorithm}
\definecolor{light-gray}{gray}{0.75}
\algrenewcommand{\algorithmiccomment}[1]{\hskip3em{{\footnotesize \textcolor{light-gray}{$\blacktriangleright$}}} #1}

\usepackage{multirow}
\usepackage{rotating}
\usepackage{booktabs}
\usepackage{colortbl}
\usepackage{tablefootnote}

\usepackage{array}
\newcolumntype{L}[1]{>{\raggedright\let\newline\\\arraybackslash\hspace{0pt}}m{#1}}
\newcolumntype{C}[1]{>{\centering\let\newline\\\arraybackslash\hspace{0pt}}m{#1}}
\newcolumntype{R}[1]{>{\raggedleft\let\newline\\\arraybackslash\hspace{0pt}}m{#1}}

\usepackage[pdftex,colorlinks=true,urlcolor=blue,citecolor=blue]{hyperref}

\usepackage{xspace}

\usepackage{enumitem}

\usepackage{blindtext}

\usepackage{listings}
\usepackage{xcolor}

\definecolor{codegreen}{rgb}{0,0.6,0}
\definecolor{codegray}{rgb}{0.5,0.5,0.5}
\definecolor{codepurple}{rgb}{0.58,0,0.82}
\definecolor{backcolour}{rgb}{0.95,0.95,0.92}

\lstdefinestyle{mystyle}{
    backgroundcolor=\color{backcolour},   
    commentstyle=\color{codegreen},
    keywordstyle=\color{magenta},
    numberstyle=\tiny\color{codegray},
    stringstyle=\color{codepurple},
    basicstyle=\ttfamily\footnotesize,
    breakatwhitespace=false,         
    breaklines=true,                 
    captionpos=b,                    
    keepspaces=true,                 
    numbers=left,                    
    numbersep=5pt,                  
    showspaces=false,                
    showstringspaces=false,
    showtabs=false,                  
    tabsize=2
}

\usepackage{orcidlink}

\begin{document}
\title{AeroGen: Structured Prompting for Low-Code Drone-as-a-Service Mission Composition~\thanks{To appear in the Proceedings of the \textit{International Conference on Service Oriented Computing (ICSOC)}, 2026}}

\author{
\normalsize
Kautuk Astu~\orcidlink{0009-0000-6268-811X} and
Yogesh Simmhan~\orcidlink{0000-0003-4140-7774}\\
\normalsize\textit{Department of Computational and Data Sciences,}\\
\normalsize\textit{Indian Institute of Science, Bangalore 560012, India}\\
\normalsize\texttt{\{kautukastu, simmhan\}@iisc.ac.in}
}

\date{}
\maketitle
\pagestyle{plain}

\begin{abstract}
Developing autonomous drone applications requires developers to compose navigation, sensing, analytics and compute services while satisfying physical and environmental constraints. Large Language Models (LLMs) can generate code from natural-language requirements, but direct prompting remains unreliable for cyber-physical \textit{Drone-as-a-Service (DaaS) missions} because incorrect API calls, unsafe flight logic or missing analytics choices can cause mission failure. We propose \textit{\agen}, an LLM-driven low-code framework that translates mission requirements into executable DaaS programs using \textit{structured guardrails}. \agen combines static guardrails, such as API signatures and flight constraints, with dynamic robot, runtime and world information so that the LLM can compose navigation, sensing, analytics and execution services in one generation attempt. We evaluate \agen on 20 navigation tasks and 5 analytical missions across simulated and real-world environments.
With frontier LLMs, \agen succeeds on all runs (25/25: $5$ analytical missions $\times$ five runs) under tested conditions. Its guardrail ablation improves analytical mission success from $8\%$ to $100\%$, with code-generation time $<200$\,s and effective cost $<$US\$\,$0.50$ per mission.
\end{abstract}

\section{Introduction}
\label{sec:intro}

Unmanned Aerial Vehicles (UAVs) are increasingly used for infrastructure inspection, agricultural monitoring, search-and-rescue and delivery~\cite{cable-inspection,farm-survey,drone-delivery}. These applications are naturally service-oriented: a mission composes navigation, sensing, analytics and compute-placement services under physical constraints imposed by the drone and its operating environment. However, developing such application-level autonomy remains difficult. A developer must \textit{translate a high-level mission requirement into executable logic} that selects sensors, invokes analytical models, schedules edge or cloud resources, and maintains safe navigation.

\textit{Drone-as-a-Service (DaaS) platforms} address part of this challenge by exposing drone capabilities as higher-level abstractions~\cite{aerodaas,aerodaas_tsc,buzz}. These abstractions reduce the need to program low-level flight-control and hardware operations. However, they do not remove the need for \textit{mission-level composition}. A developer must still decide which navigation, sensing, analytics, and compute services to invoke, in what order, and under which operational constraints. Automating this translation from mission intent to executable service composition is therefore necessary for \textit{low-code DaaS application development}.
Recent service-computing work has similarly treated drone operations and cloud applications as service-composition problems~\cite{rizvi2025droneqos,tosca}.

\begin{figure}[!t] \centering 
\begin{minipage}[t]{0.65\columnwidth} \centering 
\includegraphics[width=\linewidth]{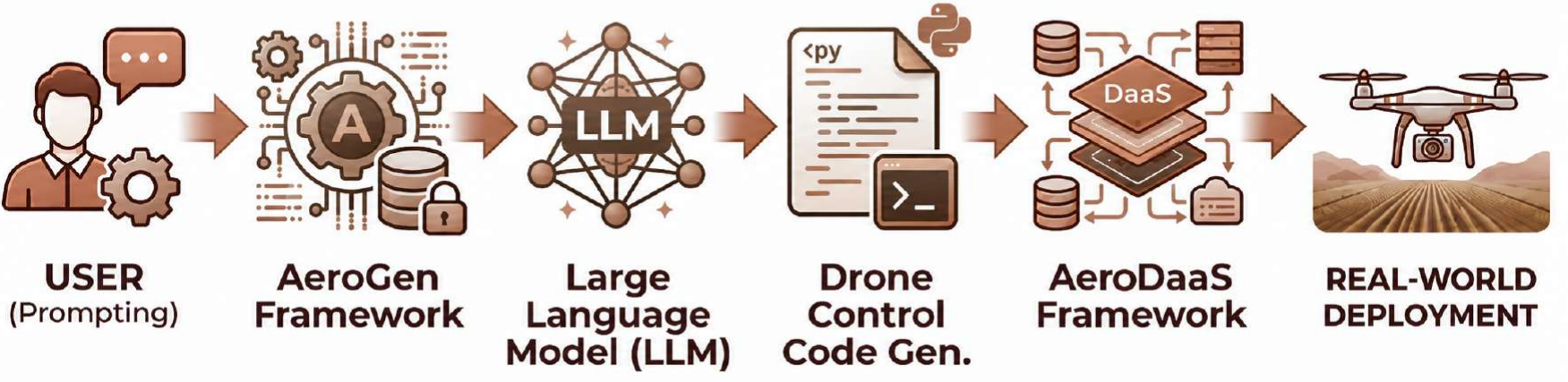} 
\caption{\agen translates natural-language mission requirements into executable DaaS programs.} \label{fig:overview} 
\end{minipage} \hfill 
\begin{minipage}[t]{0.33\columnwidth} \centering 
\includegraphics[width=\linewidth]{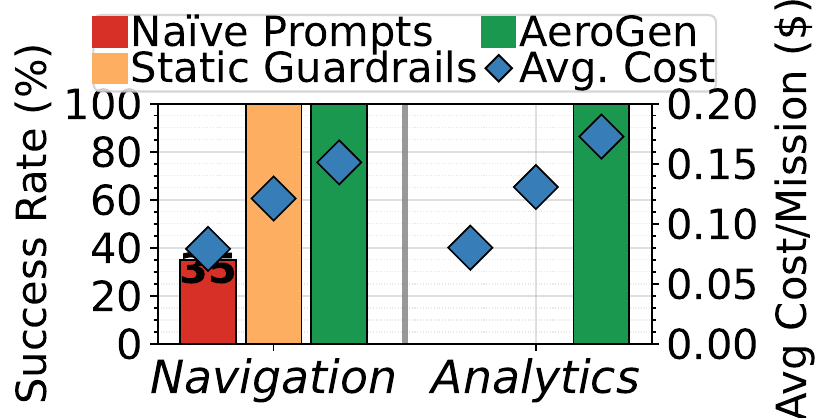} 
\caption{Success rate using Claude Opus 4.8.} \label{fig:opus-sr}
\end{minipage} 
\end{figure}

Large Language Models (LLMs) provide a promising mechanism for mission-to-program translation. Recent LLMs generate code and support planning, workflow automation and multi-agent coordination~\cite{smart-llm}, including for automated service generation and composition~\cite{tosca,peng2025mcpybarra}.
However, \textit{cyber-physical service composition} requires more than syntactically valid code. In our benchmark, na\"{i}ve code generation with \textit{Claude Opus 4.8} succeeds on \textit{only 35\%} of navigation tasks and \textit{0\%} of analytical missions (Fig.~\ref{fig:opus-sr}). These failures occur because a capable model can still generate logically incorrect trajectories, misuse service APIs, or select sensing and analytics operations that are inconsistent with the available runtime.

Prior work has improved LLM-generated drone programs using structured prompts and closed-loop feedback~\cite{gsce,code-as-service,clg-icra}. These methods improve \textit{navigation-centric} code generation, but have limited support for \textit{analytical DaaS missions} that require joint reasoning over navigation, sensing, analytics and compute runtime. \textit{Closed-loop methods}~\cite{clg-icra} improve reliability by adding repeated generation and evaluation steps, but increase Time-to-Solution (TTS) and token cost. A DaaS mission thus requires the code generator to decide not only \textit{where} the drone should fly, but also \textit{what} data it should collect, \textit{which} analytics services should process the data, and \textit{how} to compose these into one executable mission.

We propose \textit{\agen}, an LLM-driven low-code framework that generates executable DaaS mission programs from natural-language requirements (Fig.~\ref{fig:overview}). \agen combines a \textit{structured guardrail taxonomy} with our \adaas drone SDK~\cite{aerodaas}. The taxonomy combines \textit{static} programming constraints, such as API signatures, lifecycle rules, and flight constraints, with \textit{dynamic robot} information, such as battery endurance and sensors; \textit{runtime} information, such as available analytical models and compute resources; and \textit{world} information, such as obstacles, flight corridors, and operating rules. This context allows the LLM to compose navigation, sensing, analytics and execution services while generating a mission program in a single shot.

We evaluate \agen along two properties needed for practical LLM-driven DaaS programming: \textit{reliability} and \textit{efficiency}. For reliability, across the evaluated mission suite and tested deployment settings,  \textit{\agen succeeds on all single-attempt runs} with frontier LLMs (GPT-5.5, Opus 4.8)~\footnote{All rates in this paper are computed over five runs per mission and are interpreted as descriptive proportions for the tested conditions}  (\S\ref{sec:eval}). Its guardrail ablation shows that analytical mission success improves from 8\% with only static guardrails to 100\% with the complete static, robot, runtime and world context. For efficiency, \agen translates 70--102 word human-specified missions into complete executable programs while keeping code-generation TTS below $200\,s$ and effective monetary cost below US\$~0.50 per mission.

We make the following specific contributions: 
\begin{enumerate}[leftmargin=*] 
\item We introduce \textit{\agen, an LLM-driven low-code framework for DaaS mission composition}. \agen translates natural-language mission requirements into executable drone programs by combining static programming guardrails with dynamic robot, runtime, and world information (\S~\ref{sec:design}).
\item We define a \textit{DaaS mission benchmark} that extends \textit{navigation-centric} drone-code-generation tasks with five \textit{analytical missions}: multi-destination delivery, farm survey, cable inspection, radio-tower inspection, and search-and-track. These missions compose navigation, sensing, analytics, and compute services from imperative and declarative user prompts (\S~\ref{sec:apps}).
\item We \textit{evaluate \agen} across multiple LLMs and drone-programming backends. \agen achieves 100\% single-shot success on all evaluated missions with frontier LLMs, quantifies the impact of LLM capability on mission generation, improves analytical mission success from 8\% to 100\% through guardrail composition, and maintains low code-generation TTS and monetary cost (\S~\ref{sec:eval}).
\end{enumerate}

In summary, \agen does not introduce a new code-generation algorithm, but reuses structured prompting through a novel taxonomy of dynamic robot/runtime/world guardrails for cyber-physical service composition, evaluated using the DaaS mission benchmark and an empirical study of context class required by each mission type.

\section{Background and Related Work}
\label{sec:related}

\para{LLMs for Service Composition and Generation} Recent service-computing work has used Large Language Models (LLMs) to translate natural-language requirements into service artifacts. Meflah et al. generate provider-independent \textit{TOSCA} compositions from natural-language requests, identifying services and dependencies that can support later discovery, selection and execution~\cite{tosca}. MCPybarra uses a multi-agent framework to generate Model Context Protocol (MCP) services at low cost, while LLM-assisted microservice-generation approaches restructure monolithic applications into service-oriented components~\cite{peng2025mcpybarra,llm-microservices}. These systems reduce developer effort for software service construction. In contrast, \agen targets cyber-physical DaaS mission composition, where the generated program must also satisfy robot, world and runtime constraints while composing navigation, sensing, analytics and compute services.

Agentic orchestration frameworks can interleave reasoning, actions, and tool calls~\cite{react,peng2025mcpybarra}, but executable UAV missions also require physical and runtime grounding in battery limits, camera coverage, flight corridors, available analytics, and compute placement. Thus, orchestration is complementary to \agen: an agent may decide when to invoke mission generation, while \agen defines the constrained DaaS context needed to generate a valid mission program.

\para{Drone Services and DaaS Abstractions} Drone applications are increasingly modeled as services whose quality, availability and execution context affect mission outcomes. Prior work studies QoS fulfillment for Drone Services, showing that service-oriented abstractions are relevant for drone operations beyond low-level control~\cite{rizvi2025droneqos}. Drone middleware such as \textit{AeroStack2} and \textit{Buzz} provides hardware abstraction and low-level programming interfaces for UAV applications~\cite{aerostack2,buzz}. Our prior work on \textit{AeroDaaS} raises the abstraction level by exposing navigation, sensing, analytics, telemetry, and environment interfaces for Drone-as-a-Service applications~\cite{aerodaas}. \agen builds on this service abstraction, but addresses a different problem: automatically composing these services from natural-language mission intent using LLMs and structured mission context.

\para{LLM-Generated Drone and Robot Programs} LLMs have also been used for robotic planning, task decomposition and language-based control~\cite{gsce,safety-llm-driven-robotics,ImpedanceGPT,smart-llm}. \textit{GSCE} improves open-loop drone-code generation using structured prompt segments for guidelines, skill APIs, constraints, and examples~\cite{gsce}. \textit{CLG} adds a closed-loop generation--simulation--evaluation pipeline in AirSim, where trajectory observations guide code repair across repeated attempts~\cite{clg-icra,airsim2017fsr}. \textit{These systems are the closest State-of-the-Art (SOTA) LLM-based UAV code-generation baselines, but they are navigation-centric}: they primarily generate flight-control code, whereas \agen generates executable DaaS mission programs that compose navigation, sensing, analytics, and runtime compute services using dynamic robot, runtime and world context.

Other robotics-oriented LLM systems focus on safety monitoring, perception-grounded control, or multi-agent planning. Cross-layer supervision detects potential safety violations through worker and supervisor LLMs~\cite{safety-llm-driven-robotics}, \textit{ImpedanceGPT} uses vision-language models for impedance-based drone-swarm control~\cite{ImpedanceGPT}, and \textit{SMART-LLM} coordinates task allocation among multiple robots~\cite{smart-llm}. These approaches are complementary to \agen. Rather than adding runtime supervision or multi-robot coordination, \agen addresses a different problem: single-shot generation of executable cyber-physical service compositions whose correctness depends on API grounding, robot capabilities, runtime analytics, compute placement and world constraints.

\medskip
Overall, prior work establishes the usefulness of LLMs for service composition, constrained code generation, robot planning and DaaS abstractions. \agen combines these directions in three ways. \textit{First}, it treats a drone mission as an executable DaaS service composition rather than only a provider-independent workflow or a navigation script. \textit{Second}, it grounds generation using dynamic robot, runtime and world context, enabling the LLM to jointly select flight actions, sensors, analytics and compute resources. \textit{Third}, it evaluates the generated programs through end-to-end mission execution across navigation, analytical, simulated and real-world missions, rather than only plan validity or code-generation accuracy.

\section{\agen Design}
\label{sec:design}

\begin{figure}[t]
\centering
\includegraphics[width=0.6\columnwidth]{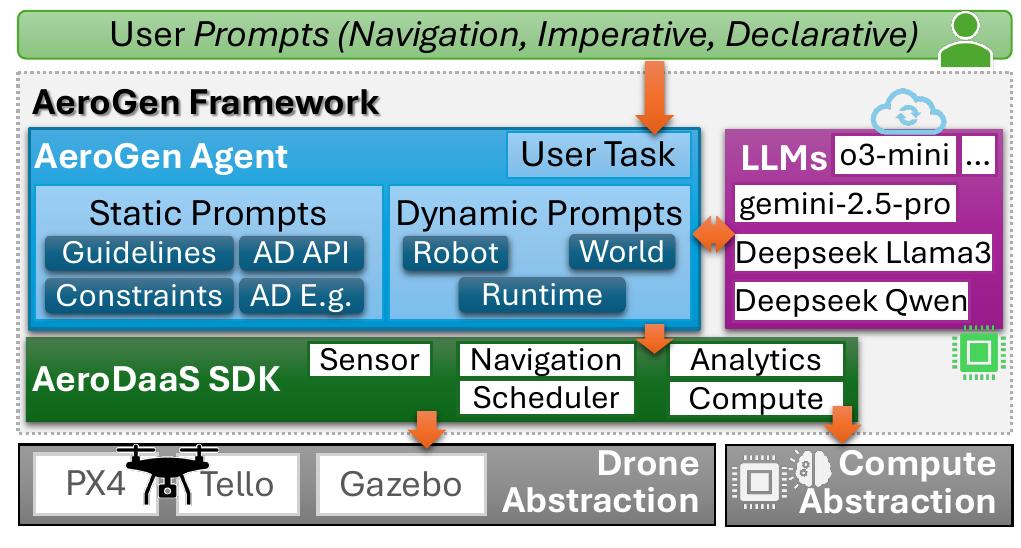}
\caption{\agen architecture. The Agent builds guardrails, the LLM generates mission code and the DaaS backend executes it.}
\label{fig:architecture}
\end{figure}

\agen treats autonomous drone programming as constrained service composition. A mission is expressed as a composition of navigation, sensing, analytics and scheduling operations, rather than as low-level flight-controller commands. This abstraction allows the LLM to reason over service intent and operational constraints while avoiding hardware-specific command details. \textit{The key design principle is to move correctness guidance into the prompt context.} The LLM receives \textit{structured information} about service APIs, physical constraints, robot capabilities, runtime analytics and world rules before generating code. \agen uses our \adaas~\cite{aerodaas} as the default DaaS backend and also supports \as~\cite{aerostack2} and AirSim~\cite{airsim2017fsr}. We evaluate the backend tradeoffs in \S~\ref{subsec:results:analytical}.

We formulate \textit{mission generation} as $P = f(M, A, C)$, where $M$ is the natural-language mission request from the user, $A$ is the set of DaaS platform API descriptions given to the LLM, $C$ is the structured context of static and dynamic constraints  defined by \agen and used to ground the generated service composition, and $P$ is the generated mission program.

\subsection{\agen Architecture}
\label{sec:arch}

The \agen runtime has three components: the \textit{\agen Agent}, the \textit{LLM Cognitive Core} and our \textit{\adaas SDK/Backend}. Together, they translate a natural-language mission into executable DaaS code and execute it over a simulator or a physical drone platform.

The \textbf{\agen Agent} constructs the \textit{system prompt} that bounds the LLM's code-generation task. It first retrieves \textit{Static Guardrails}, which define invariant information such as the coordinate frame, API signatures, lifecycle rules, scheduling guidelines and flight constraints. It then adds \textit{Dynamic Guardrails} for the current mission, including \textit{Robot}, \textit{Runtime} and \textit{World} information. These segments inform the LLM about the drone's capabilities, available analytics and compute resources, and feasible operating environment.

The \textbf{LLM Cognitive Core} interprets the user mission and generates Python code using the DaaS APIs. \agen is \textit{model-agnostic}: our evaluation studies frontier LLMs and lower-cost models to quantify how model capability affects mission success, TTS and generation cost. This separation keeps the prompt-construction logic independent of the specific LLM backend.

The \textbf{\adaas SDK/Backend} executes the generated Python mission program through high-level DaaS interfaces. \texttt{IEnvironment} exposes robots, sensors, and compute resources; \texttt{IRobot} handles navigation; \texttt{ISensor} streams sensor data; \texttt{IAnalyse} deploys and executes analytical models on edge or cloud resources; \texttt{INavigate} maps waypoints or analytics outputs to navigation actions; and \texttt{IScheduler} manages task priorities. These interfaces hide low-level flight-control and hardware details, allowing the LLM to compose mission services rather than directly issue controller commands.

For validation, the \agen Agent can first execute the generated program with \adaas in \textit{Gazebo}. After execution, it extracts telemetry logs and compares the trajectory against a task-specific ground truth to determine mission success. In our evaluation, we report \textit{single-shot success} for the first generated program; we do not perform any repair or regeneration attempt. Once validated, the same DaaS program can target the physical drone and compute hardware.

\subsection{Guardrail Prompts for Physical Grounding}
\label{sec:prompt-schema}
\begin{figure}[t]
    \centering
    \includegraphics[width=0.85\columnwidth]{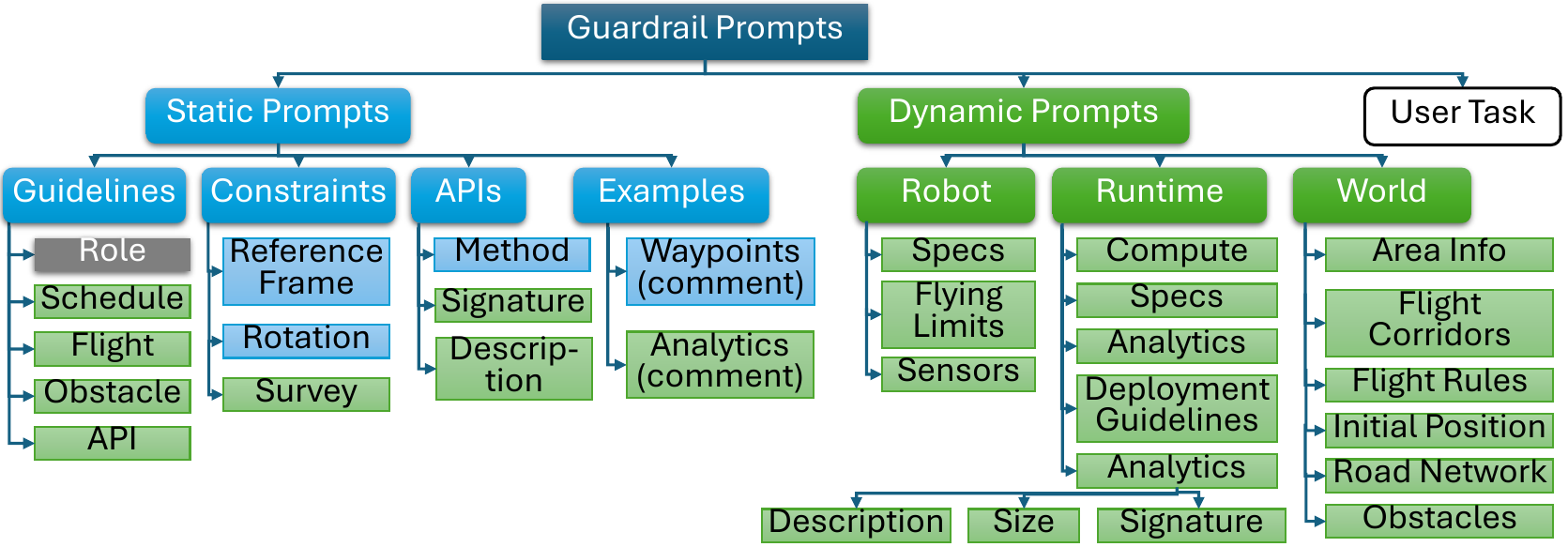}
\caption{\agen guardrail prompt schema. Grey boxes reuse prior \CLG prompt classes~\cite{clg-icra}, blue boxes modify prior classes for DaaS code generation, and green boxes add dynamic robot, runtime, and world context.}    \label{fig:prompt-schema}
\end{figure}

The guardrail schema grounds the LLM in the APIs, constraints and mission context needed to generate executable DaaS programs (Fig.~\ref{fig:prompt-schema}). Compared to \CLG~\cite{clg-icra}, \agen retains static prompt classes for guidelines, constraints, APIs and examples, but adds dynamic \textit{Robot}, \textit{Runtime} and \textit{World} information. These additions allow the generated program to account for platform capabilities, available analytics, compute resources and feasible flight behavior.

\subsubsection{Static Guardrail Prompt}
Static guardrails remain fixed across missions, UAVs and worlds. They define the permissible structure of generated programs through $4$ prompt classes: \textit{Guidelines} specify the LLM's coding role, lifecycle order, scheduling strategy, flight reference, obstacle-avoidance behavior and API-usage rules; \textit{Constraints} define operational bounds such as the coordinate frame, rotation conventions, altitude limits and survey geometry; \textit{APIs} provide DaaS interface signatures and descriptions; and \textit{Examples} provide code snippets that demonstrate valid mission structure and reduce ambiguity in API usage.

\subsubsection{Dynamic Guardrail Prompt}

Dynamic guardrails provide mission-specific context that changes with the robot, runtime and world. \textit{Robot Information} describes platform capabilities such as speed limits, battery endurance and active sensors. \textit{Runtime Information} describes available edge and/or cloud resources and the analytical models that can run on them, including model size, latency, memory needs and preferred model--resource mappings. \textit{World Information} describes the operating environment, including permitted flight corridors, obstacles, altitude limits, velocity limits and spatial rules. 
Together, these guardrails allow the LLM to select feasible analytics, route through valid flight paths, and generate mission programs consistent with the current deployment context.
These guardrail information is maintained by the platform and supplied as structured, reusable guardrails. Encoding the same information directly in every task prompt would substantially increase prompt length and require users to understand platform capabilities, available analytics, compute resources, and environmental constraints, undermining the low-code abstraction and increasing user effort.

\medskip
Besides these, the \textit{user task} discussed next specifies the mission objective in natural language. The preceding guardrails define the valid service APIs and operational context, while the LLM synthesizes the control and service-composition logic needed to execute the mission.

\section{Mission Taxonomy and Application Suite}
\label{sec:apps}

\subsection{Mission Taxonomy}
\label{sec:taxonomy}

We evaluate \agen using a three-level mission taxonomy that increases the required service-composition complexity. \textit{Navigation tasks} require only motion planning and flight-control API usage. \textit{Imperative analytical missions} specify both the mission goal and the main procedural steps, requiring the generated program to compose navigation, sensing, and analytics services. \textit{Declarative analytical missions} specify the goal, requiring the LLM to infer search strategy, sensing behavior, analytics selection and task-switching logic. This taxonomy separates trajectory generation from end-to-end DaaS mission composition.

The navigation tier reuses the 20 advanced drone-code-generation tasks from prior LLM-based drone-control work~\cite{gsce,clg-icra}. These tasks test waypoint traversal and path planning under structured world rules, but do not require sensing or analytics. The imperative tier adds analytical missions in which the user specifies the goal and the main steps. The declarative tier further removes those procedural hints, so the generated program must derive the mission strategy and compose the required services from the guardrail context. The latter two tiers are the primary tests of \agen's ability to generate DaaS mission programs rather than only navigation code.

\begin{table}[t!]
\centering
\setlength{\tabcolsep}{1pt}
\renewcommand{\arraystretch}{1}
\def\thickhline{\noalign{\hrule height1pt}}
\caption{\hlyy{Imperative (yellow cells)} and \hlrr{declarative (red)} missions and runtime config. in \hlb{simulated (blue)} and \hlg{real (green)} worlds.}
\label{tab:workloads}
\scriptsize
\begin{tabular}{L{1.35cm}||C{1.95cm}||C{2.45cm}|c|c|c|c|C{1.2cm}}
\hline
\textbf{Mission}  & \textbf{World Setup} & \textbf{Analytics} & \textbf{FPS} & \textbf{Duration} & \textbf{Area/Path} & \textbf{Alti.} & \textbf{Speed} \\
\thickhline
\cellcolor{yellow!40}  & \cellcolor{blue!15}Gazebo small city \textit{(Sim)} & -- & 30 & 6 min & $600\ m$ & 10 m & 2 m/s \\
\cline{2-8}

\cellcolor{yellow!40}\multirow{-3}{1.15cm}{Multi-destination delivery} & \cellcolor{green!20}Outdoor field \textit{(Real)} & -- & 5 & 7 min & 42 m & 2 m & 0.25 m/s \\
\hline

\cellcolor{yellow!40} & \cellcolor{blue!15}Gazebo small city \textit{(Sim)} & Video logger & 30 & 2 min & 900 $m^2$ & 3 m & 2 m/s \\
\cline{2-8}

\cellcolor{yellow!40}\multirow{-3}{1.15cm}{Farm survey}& \cellcolor{green!20}Outdoor field \textit{(Real)} & Video logger & 5 & 4.5 min & 100 $m^2$ & 1.5 m & 0.25 m/s \\
\hline

\cellcolor{red!20}Cable inspection & \cellcolor{blue!15}Gazebo small city, 1 radio tower \textit{(Sim)} & YOLOv11x (cable detection) & 3 & 9.5 min & 100 m & 44 m & 0.1 m/s -- 0.5 m/s \\
\thickhline

\cellcolor{red!20}Radio Tower Inspection & \cellcolor{blue!15}Gazebo small city, 3 radio towers \textit{(Sim)} & OpenCV ORB Image Match & 3 & 57 min & 43,200 $m^2$ & 45 m & 2 m/s \\
\hline

\cellcolor{red!20}Search \& Track & \cellcolor{green!20}Outdoor field \textit{(Real)} & YOLOv11x (hazard vest detect) & 5 & 4 min & 25 m & 1.5 m & 0.1 m/s -- 1.0 m/s \\
\hline

\end{tabular}
\end{table}

In addition to these navigation tasks, we define five analytics-driven DaaS missions that combine navigation, sensing, analytics, and runtime compute decisions (Table~\ref{tab:workloads}). The suite includes two imperative missions and three declarative missions, spanning simulated urban environments and real-world field deployments. We use these missions to evaluate single-shot mission success, guardrail ablations, backend choice and LLM capability in \S~\ref{sec:eval}.

\subsection{Analytical Mission Suite}

\textbf{Multi-destination delivery} is an \textit{imperative mission} in which the drone visits three delivery locations, descends to drop-off a virtual package at each and returns to the depot before violating battery or flight-corridor constraints. This mission mainly tests waypoint sequencing, altitude control, obstacle avoidance and battery-aware route completion.

The task prompt to \agen is: \hly{\textit{\small``You are performing a multi-destination package delivery mission starting from a depot. Navigate to delivery locations I22, I34 and I43 while avoiding obstacles. At each delivery location, descend to 1 meter and hover for 10 seconds to complete the virtual delivery. Then proceed to make next delivery. After all 3 deliveries, return back to the starting location to complete the mission. Refer to the  mission\_operational\_information for area information and make sure to avoid obstacles. Abort deliveries and return to depot if you are going to run out of battery.''}}.

\textbf{Farm survey} is an \textit{imperative mission} in which the drone records video while traversing the boundary of a rectangular field and orienting its camera toward the farm interior. This mission tests the composition of geometric navigation, continuous sensing, and camera-orientation constraints in both simulation and real-world deployments. \textit{(The user prompt for this and next two missions is omitted for brevity.)}

\textbf{Cable inspection} is a \textit{declarative mission} in which the drone follows high-tension cable infrastructure while recording video and invoking a cable-detection model. This mission tests whether \agen can combine ascent, target acquisition, continuous sensing, analytics deployment, and analytics-guided navigation.

\textbf{Radio-tower inspection} is a declarative mission in which the drone searches a large simulated area, detects towers, performs tower-specific inspection when a tower is found, and resumes search until the region is covered or battery constraints require return. This mission tests long-horizon planning, search-to-inspection mode switching, and analytics-driven navigation.

\textbf{Search-and-track} is a \textit{declarative real-world mission} in which the drone searches a bounded area for a person wearing a hazard vest and switches from coverage search to tracking after detection. This mission tests sensing, person-detection analytics, task interruption, and reactive mission reconfiguration. 

The user prompt is: \hly{\textit{\small ``You are supposed to perform search and rescue mission. The search need to be done in an area of 8m * 4m. The drone should systematically search the area to cover the entire 8m*4m search region. Simultaneously, the drone should also look for a person wearing hazard vest, upon detecting the person, the drone should abort it's search mission and immediately transition from search to tracking the person.''}}.

Rather than exhaustively covering all DaaS deployments, our benchmark isolates key sources of mission-generation complexity: sensing, analytics selection, runtime compute choice, world constraints, and task switching. The five analytical missions therefore serve as representative stress cases for evaluating service-composition behavior, not as a complete taxonomy of drone applications.

\section{Evaluation}
\label{sec:eval}
We evaluate \agen along three questions: (i) \textit{Do structured guardrails improve single-shot reliability for navigation and analytical DaaS missions?}, (ii) \textit{Are the generated programs efficient in TTS and monetary cost?}, and (iii) \textit{How do backend and LLM choices affect mission success?}. We first describe the experimental setup and metrics, then compare \agen against navigation-centric baselines and analytical mission baselines.

\begin{figure}[t]
\centering
\subfloat[Gazebo Small City, Farm \& Towers]{%
    \includegraphics[width=0.24\columnwidth]{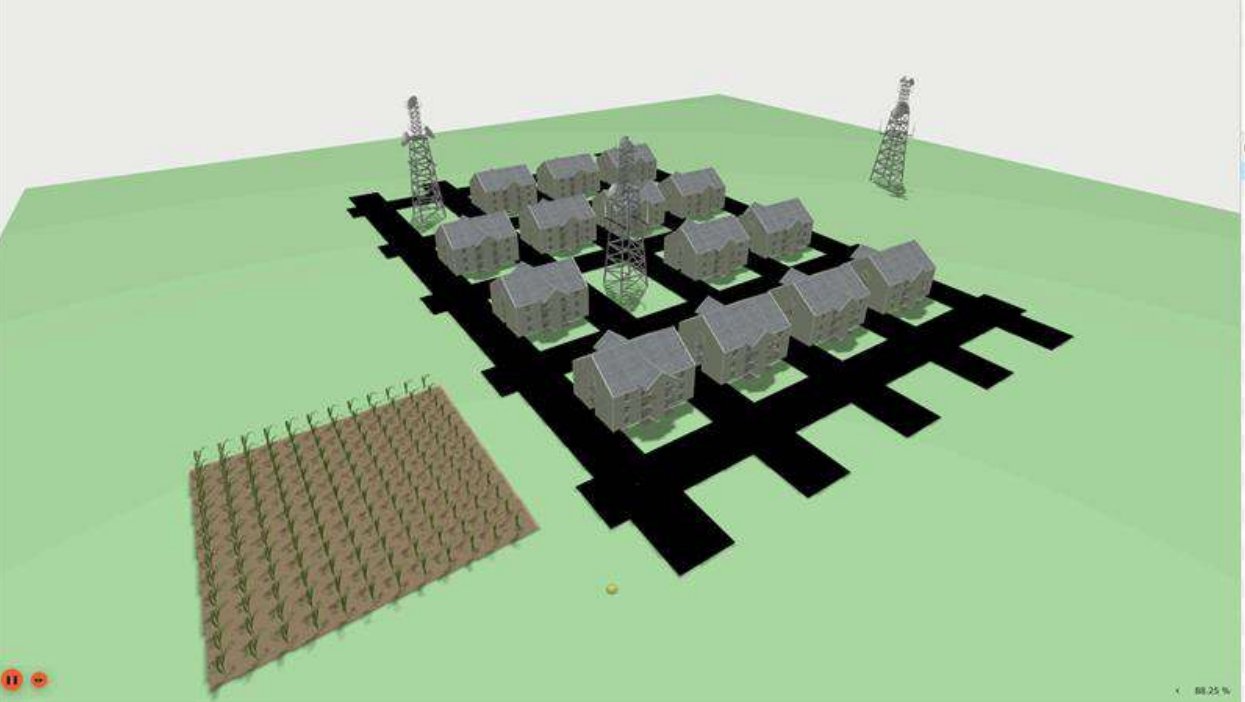}%
    \label{fig:gazebo-city}%
}~
\subfloat[Search and track in real-world]{%
    \includegraphics[width=0.24\columnwidth]{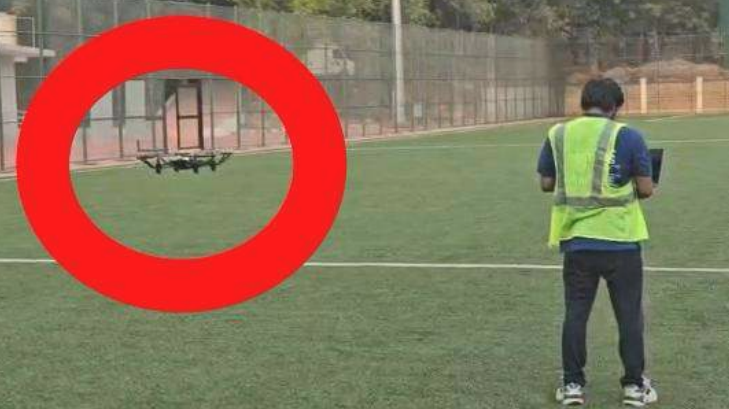}%
    \label{fig:exp-search-rescue}%
}~
\subfloat[Farm survey in real-world (side)]{
    \includegraphics[width=0.24\columnwidth]{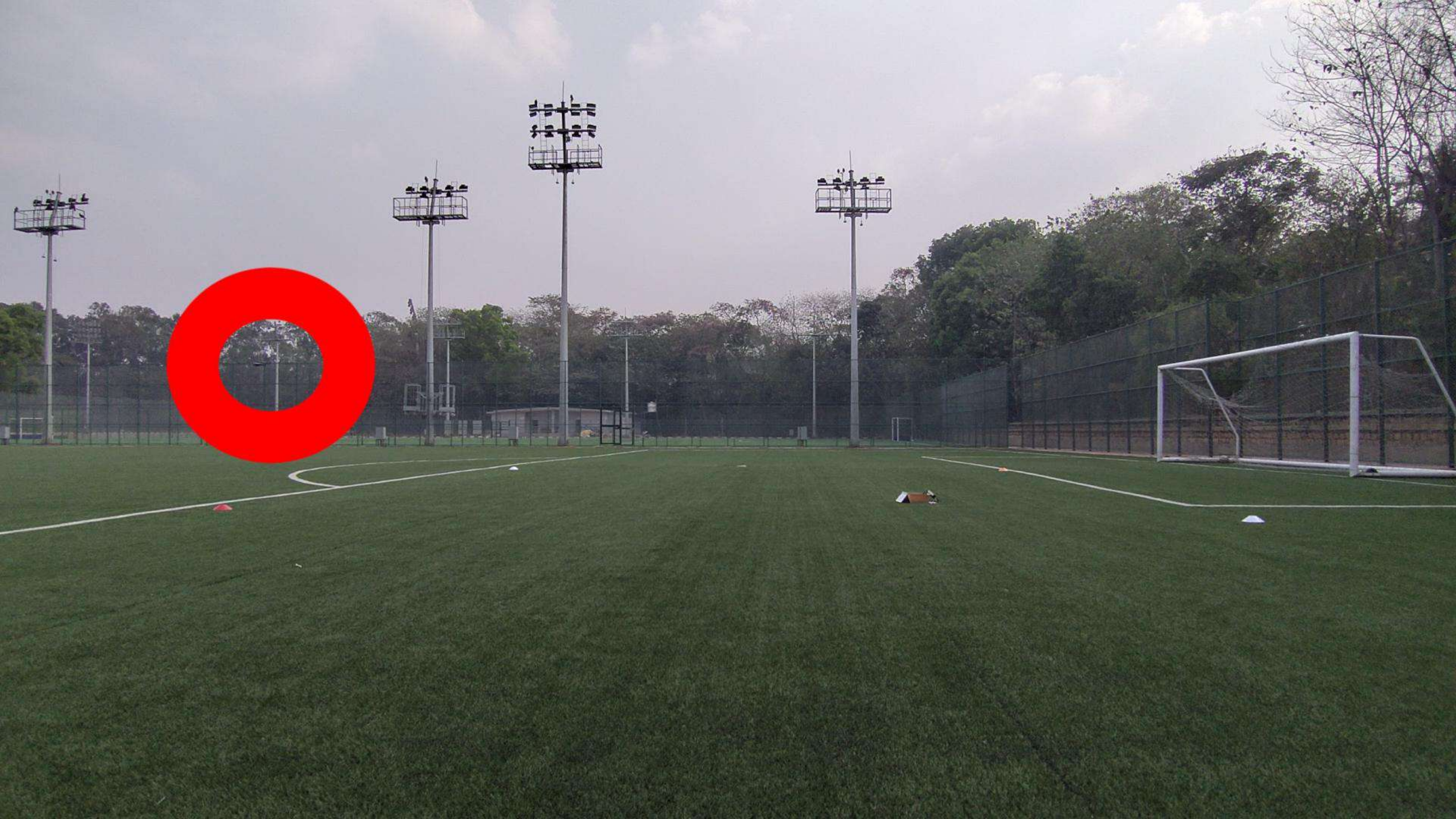}
    \label{fig:exp-farm-survey-side}
}~
\subfloat[Farm survey in real-world (top)]{
    \includegraphics[width=0.24\columnwidth]{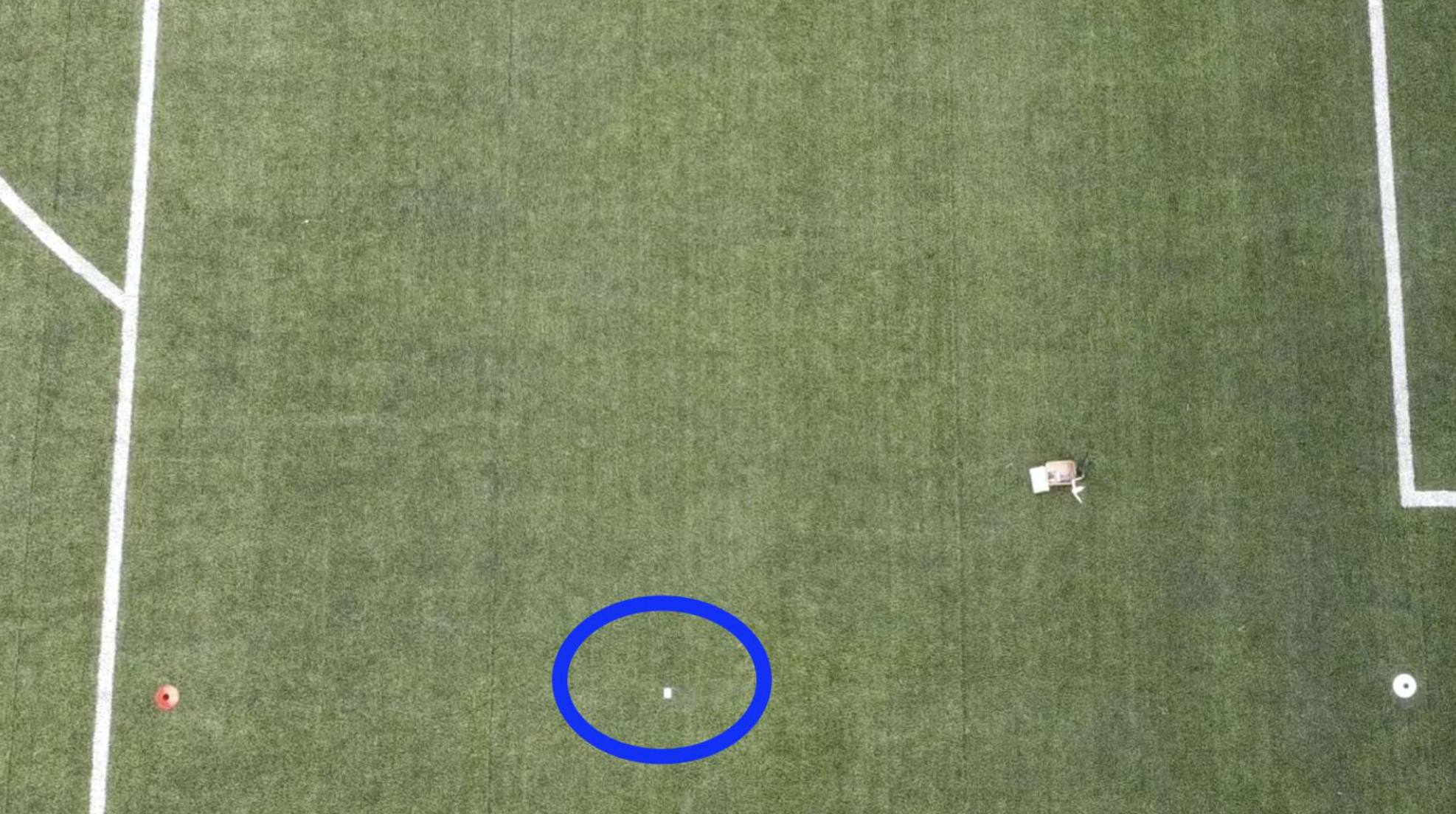}
    \label{fig:exp-farm-survey-top}
}
\caption{Experimental setup and mission execution.}
\label{fig:exp-drone-platform}
\end{figure}

\subsection{Experimental Setup}
\label{subsec:exp-setup}

\para{Runtime environment} We evaluate \agen in simulation and real-world deployments. The simulation uses \textit{Gazebo Ignition Fortress} with a \textit{custom city} containing roads, a farm, power cables and radio towers. The real-world tests use a \textit{DJI Ryze Tello drone} with a \textit{Jetson Orin Nano edge node} on a campus field. Unless otherwise stated, \agen uses AeroDaaS v1.5~\cite{aerodaas} and GPT-5.5~\cite{gpt-5-5}.

In the absence of an analytics-capable baseline, we use \CLG only as a navigation reference. CLG cannot compose analytics services by construction; hence, its analytical results measure a functionality gap rather than comparative generation quality. The guardrail ablation (\S\ref{sec:ablation}) is the controlled comparison.

\para{Success criterion} For navigation and imperative analytical missions, the \textit{user prompt} fixes the target waypoint sequence and altitude profile in their task. We compare the executed telemetry trajectory $T_{exec}$ with the expected trajectory $T_{gt}$ and mark a run successful when corresponding trajectory points remain within $0.2\,m$ while satisfying altitude and boundary constraints.
We treat the $0.2\,m$ tolerance as a tunable evaluation hyperparameter rather than a universal safety bound.
This primarily tests instruction-following and correct API grounding.
For \textit{declarative missions}, multiple valid trajectories may exist. A run succeeds if telemetry satisfies mission-specific predicates: required coverage or target inspection, sensing/analytics invocation, behavior transition after detection, safe termination or return before battery violation, and no altitude, boundary or obstacle-rule violation; this tests the autonomous strategy synthesis.

\para{Metrics} We report \textit{Success Rate}, the fraction of successful runs across five independent generation attempts per mission and configuration; \textit{Token Usage}, including prompt, cached, reasoning, and output tokens; and \textit{Time-to-Solution (TTS)}, measured for code generation alone and end-to-end with simulation execution. Since one success changes a five-run mission rate by 20 percentage points, we interpret results as empirical proportions with Wilson intervals, not reliability guarantees. We report user-prompt lines/words to quantify input effort.

\subsection{Performance on Navigation Missions}

\begin{figure}[t] \centering 
\subfloat[Navigation reliability]{
\includegraphics[width=0.25\columnwidth]{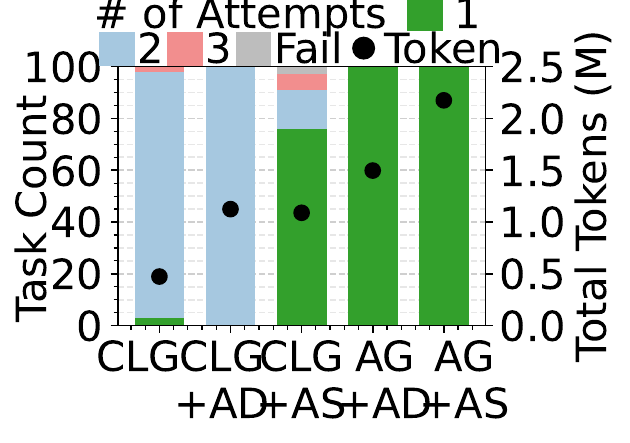}%
\label{fig:nav-attempt-stacked}%
}
\subfloat[Analytical mission reliability]{%
\includegraphics[width=0.5\columnwidth]{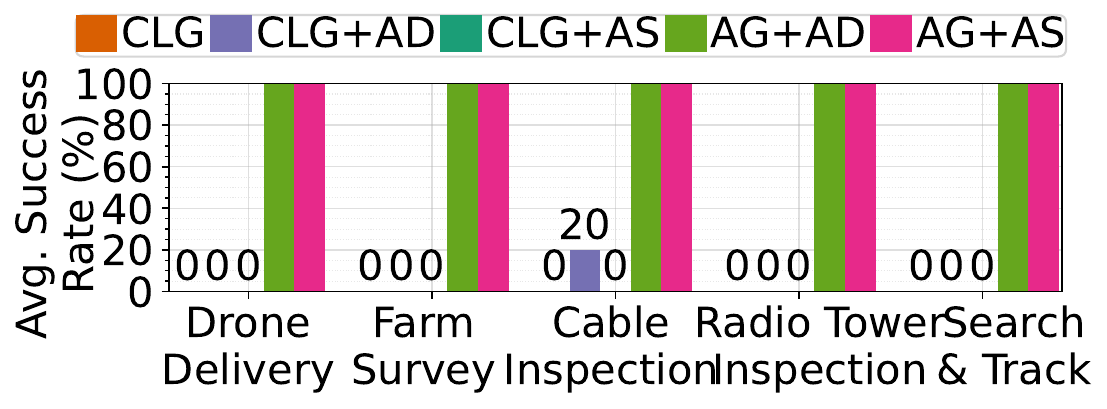}%
\label{fig:clg-vs-agen-analytical-missions}%
}%
\subfloat[DaaS efficiency]{%
\includegraphics[width=0.24\columnwidth]{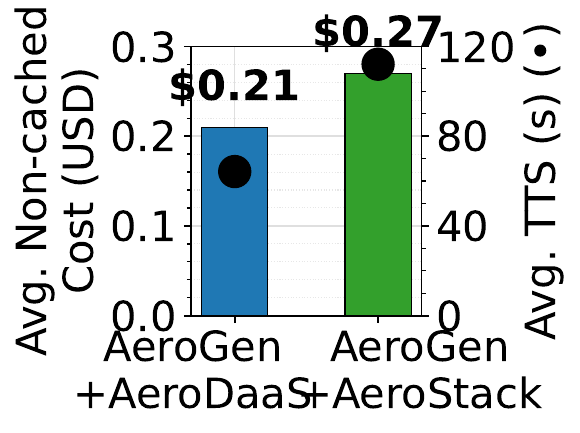}%
\label{fig:ad-vs-as-cost-tts}%
}
\caption{Reliability and backend efficiency of mission generation.}
\end{figure}

\takeaway{\agen achieves single-shot navigation success, while CLG often needs repeated attempts.} Fig.~\ref{fig:nav-attempt-stacked} compares attempts-to-success across 20 navigation tasks, each repeated five times. Native \CLG rarely succeeds in one attempt and requires up to three attempts for several tasks. Integrating \CLG with \adaas (AD) or \as (AS) reduces the number of attempts for many tasks, but does not eliminate failures. The remaining failures are caused by incorrect relative-waypoint interpretation, missing default takeoff-altitude information, evaluator errors or invalid yaw generation.

In contrast, \agen (AG) completes all 20 navigation tasks in one generation attempt with both \adaas and \as (Fig.~\ref{fig:nav-attempt-stacked}). This result shows that the static guardrails and backend-specific API grounding are sufficient to remove the repeated generation attempts required by \CLG for navigation-only workloads. It also establishes navigation as the lower-complexity case before evaluating analytical DaaS missions.

\subsection{Performance on Analytical Missions}
\label{subsec:results:analytical}

\takeaway{\agen completes every evaluated analytical run in one attempt, while navigation-centric \CLG fails on analytical missions as expected.}
With either \adaas or \as, our \agen achieves 100\% success with a 95\% Wilson CI of [0.56,1] across all five analytical missions (Fig.~\ref{fig:clg-vs-agen-analytical-missions}). In contrast, native \CLG and \CLG+AS fail on every analytical mission, while \CLG+AD succeeds only once for Cable Inspection. This isolated success indicates that higher-level DaaS APIs can sometimes help \CLG generate a valid analytical program, but API access alone does not provide the runtime, robot and world context needed for reliable mission composition. \agen closes this gap by exposing capabilities and constraints through dynamic guardrails, allowing reliable composition of navigation, sensing, analytics and execution services, rather than only for navigation tasks.

\begin{figure}[!t]
\centering
\subfloat[Aggregate success]{
    \includegraphics[width=0.33\columnwidth]{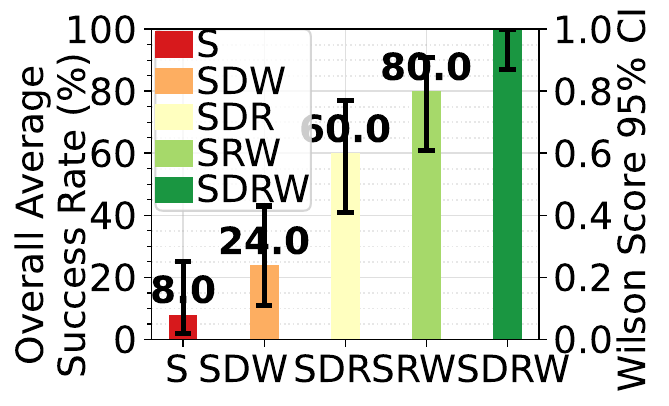}
    \label{fig:ablation-summary-gpt5}
}~
\subfloat[Per-mission success]{
    \includegraphics[width=0.6\columnwidth]{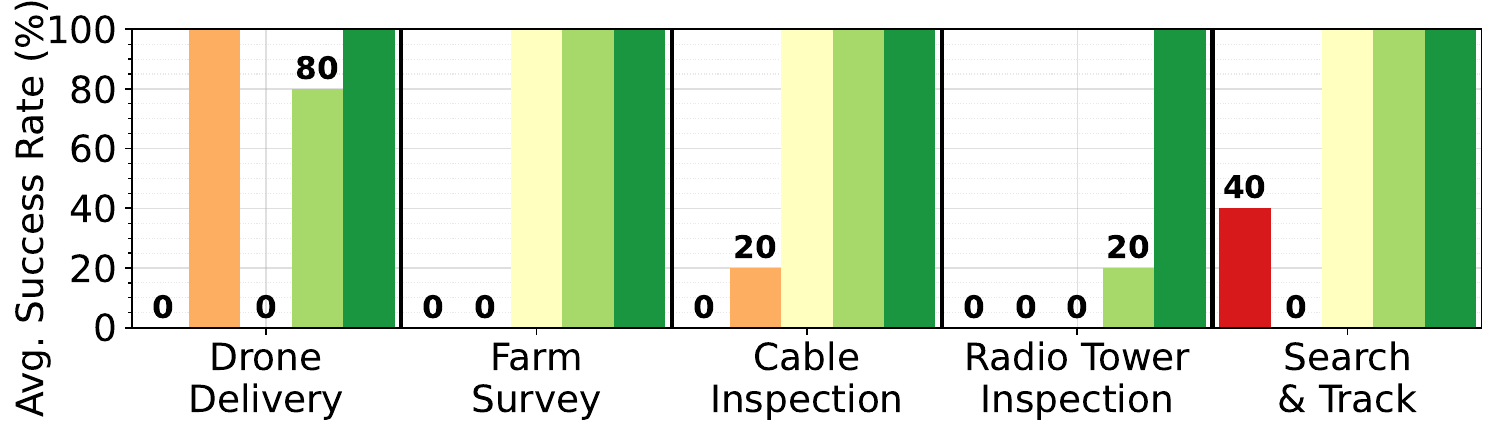}
    \label{fig:ablation-per-mission-gpt5}
}
\caption{
Guardrail ablation for analytical missions. S=\underline{S}tatic guardrails, D=\underline{D}rone/robot info., R=\underline{R}untime info., and W=\underline{W}orld info.}
\label{fig:guardrail-ablation-study}
\end{figure}

\takeaway{\adaas provides a more efficient target backend for \agen.} Although both \adaas and \as give 100\% analytical mission success for \agen with frontier LLMs, they differ in generation efficiency (Fig.~\ref{fig:ad-vs-as-cost-tts}). \agen+AD reduces \textit{average code-generation TTS} from $112\,s$ to $64\,s$ per mission and \textit{average non-cached generation cost} from \$\,$0.27$ to \$\,$0.20$ compared to \agen+AS. The backend difference is larger for lower-capability LLMs: with o3-mini, \agen+AS fails on 52\% of analytical mission attempts, while \agen+AD fails on only 35\%.
This result should be interpreted as a backend-abstraction effect rather than a universal property of \agen: more compact DaaS APIs can reduce generated-code size and the LLM's reasoning burden, but other SDKs may expose different tradeoffs.
E.g., AD generates just 80 Lines of Code (LOC) vs 344 for AS in Cable Inspection (Fig.~\ref{fig:loc-prompt-time}).
We therefore use \adaas as the default backend for \agen in the remaining experiments.

\subsection{Ablation Study of \agen's Structured Prompt Taxonomy}
\label{sec:ablation}

\takeaway{Complete guardrail context is necessary for reliable analytical missions.} We evaluate the contribution of each guardrail class by varying the information provided to the LLM for analytical mission generation. The \underline{s}tatic-only configuration (S) includes guidelines, constraints, APIs and examples. The remaining configurations add combinations of \underline{d}ynamic robot information (D), \underline{r}untime information (R), and \underline{w}orld information (W). As shown in Fig.~\ref{fig:ablation-summary-gpt5}, static guardrails alone achieve only 8\% average mission success with a $95\%$ CI of $[0.02,0.25]$. Adding dynamic context improves success to $24\%$ with SDW, $60\%$ with SDR, and $80\%$ with SRW, while the complete SDRW configuration of \agen reaches 100\% with corresponding $95\%$ CIs of $[0.11,0.43]$, $[0.41,0.77]$, $[0.61,0.91]$ and $[0.87,1.00]$, respectively.
These results show that deployment-specific robot, runtime and world information complements static API and constraint descriptions for the evaluated analytical missions. This ablation isolates the contribution of these information classes, rather than their prompt organization. Each dynamic guardrail also increases the input context; we quantify this overhead in \S~\ref{subsec:agen-overhead}.

\takeaway{Runtime and world context provide the largest mission-level gains.} Fig.~\ref{fig:ablation-per-mission-gpt5} shows that the benefit of each guardrail depends on the mission. \textbf{SDW} succeeds consistently for Drone Delivery because the mission depends primarily on navigation and does not require analytics. In contrast, Farm Survey, Cable Inspection and Search-and-Track depend on runtime information to select the correct sensing and analytics services, where SDW does not assist. \textbf{SDR} succeeds on such analytics-driven missions but fails on navigation-critical missions (e.g., Drone Delivery) when world information is missing, because the generated programs cannot reliably avoid obstacles or obey spatial constraints. \textbf{SRW} performs better overall, but Radio Tower Inspection remains sensitive to missing robot information because camera coverage and altitude determine whether the search pattern observes the towers. 
Overall, runtime information gives the largest gains for analytics-heavy missions, while world information gives the largest gains for spatially constrained missions.
We therefore use the complete SDRW guardrail configuration unless otherwise stated.
 
\takeaway{Examples provide structural scaffolding but do not replace dynamic guardrails.} Table~\ref{tbl:eg-vs-success} varies the examples included in the static prompt and their input-token cost. Removing all examples (`None') causes every analytical mission to fail, while the two waypoint examples add 2.3K input tokens but succeed only for Delivery and Survey, which resemble waypoint traversal. The single 820-token analytical example is insufficient, indicating that the LLM requires representative structural patterns for navigation and analytics. The full 3.2K-token example set succeeds across all missions, showing that examples complement static and dynamic guardrails by reducing API ambiguity and invalid mission structure.

\subsection{\agen Overhead}
\label{subsec:agen-overhead}

\begin{figure}[!t]
\centering
\subfloat[Prompt, LOC and TTS]{
    \includegraphics[width=0.5\columnwidth]{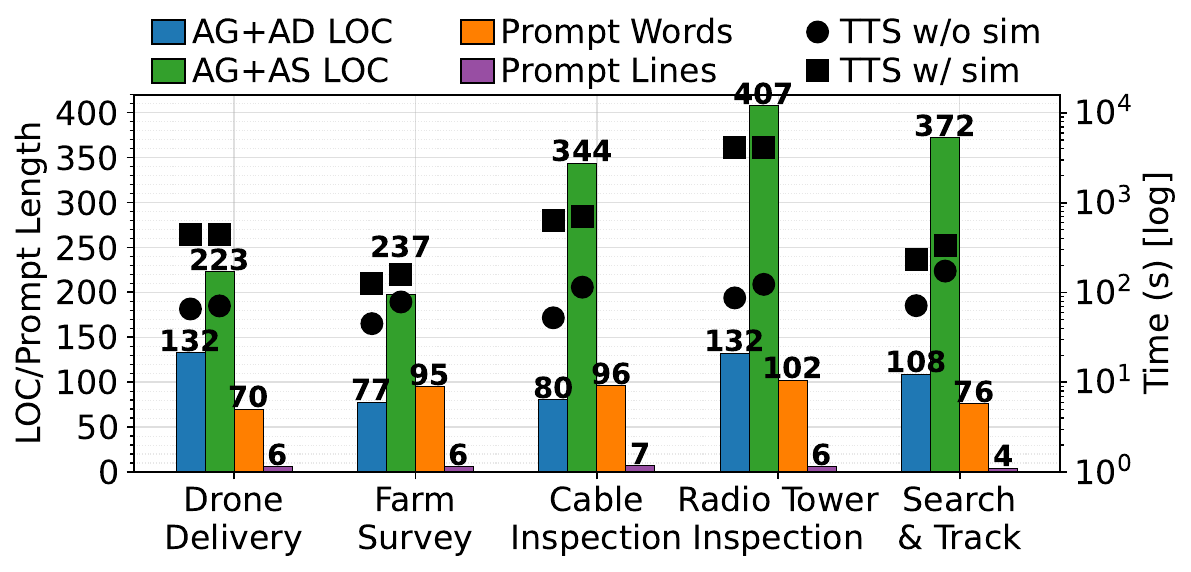}
    \label{fig:loc-prompt-time}
}~
\subfloat[Tokens, cost and prefix caching]{
    \includegraphics[width=0.5\columnwidth]{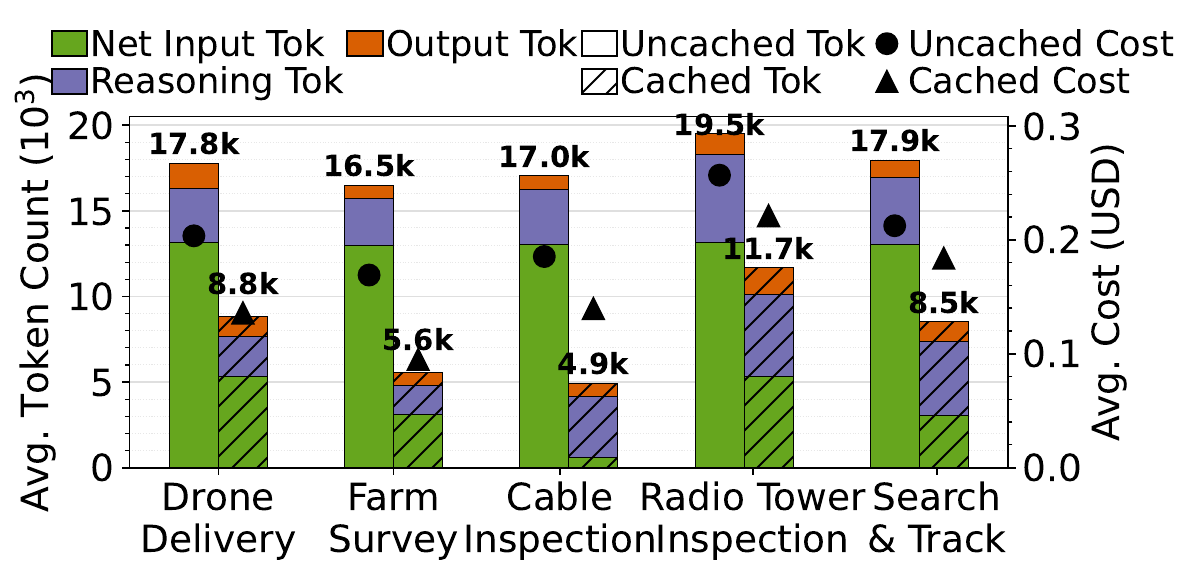}
    \label{fig:ad-ag-token-cost}
}
\caption{\agen overhead and developer effort.}
\label{fig:agen-human-comparison}
\end{figure}

\takeaway{\agen reduces developer-written code while keeping generation latency modest.} Fig.~\ref{fig:loc-prompt-time} compares the length of the user mission definition with the generated program size and code-generation TTS. Users specify the analytical missions using only 4--7 \textit{lines} of natural-language instructions containing 70--102 \textit{words}, while \agen generates complete executable programs with 70--407 \textit{LOC}. This shifts developer effort from manually implementing navigation, sensing, analytics and scheduling logic, which requires domain expertise, to specifying mission objectives and constraints. Across analytical missions, code-generation TTS stays $<200\,s$, while simulation-inclusive TTS is dominated by simulation time rather than code generation. This distinction is important because manually written drone programs typically pass simulation validation before physical deployment.

\takeaway{Prefix caching amortizes the cost of structured guardrails.} Fig.~\ref{fig:ad-ag-token-cost} shows that input and reasoning tokens dominate generation cost, while output tokens form a smaller fraction. The structured guardrails increase prompt size because they include API descriptions, flight constraints, robot information, runtime analytics and world context. However, much of this context is reusable across missions and can be cached as a prefix. With prefix caching, \agen reduces the effective average generation cost from \$\,$0.21$ without caching to \$\,$0.15$, a 29\% reduction, and keeps the effective cost between \$\,$0.09$ and \$\,$0.22$ across the evaluated missions. Thus, \agen adds domain context to improve reliability without making monetary cost proportional to the full guardrail size.


\subsection{Impact of LLM Choice on \agen}
\label{subsec:llm-ablation}

\begin{figure*}[t] \centering 
\begin{minipage}[b]{0.4\textwidth} \centering \scriptsize 
\setlength{\tabcolsep}{1pt}
\renewcommand{\arraystretch}{0.7}
\begin{tabular}{l|cccc} \hline 
\bf Mission & \bf None & \bf 2x WP & \bf 1x Anlt. & \bf All 3\\ \hline 
Delivery & \czero & \cfour & \czero & \cfour\\ 
Survey & \czero & \cfour & \czero & \cfour\\ 
Cable & \czero & \czero & \czero & \cfour\\ 
Tower & \czero & \czero & \czero & \cfour\\ 
Search & \czero & \czero & \czero & \cfour\\ \hline 
\multicolumn{5}{c}{\scriptsize 
\shortstack{
\textit{\# Examples:} \textbf{None} (0 Input tokens).\\[-0.5ex] 
\textbf{2x Waypoint} (2.3k tok.)\\[-0.5ex] 
\textbf{1x Analytical} (820 tok.).\\[-0.5ex] 
\textbf{All 3} (3.2k tok.)}}
\end{tabular}
\captionof{table}{Effect of prompt examples on mission success.} \label{tbl:eg-vs-success} 
\end{minipage}~
\begin{minipage}[b]{0.58\textwidth} 
\centering 
\includegraphics[width=\textwidth]{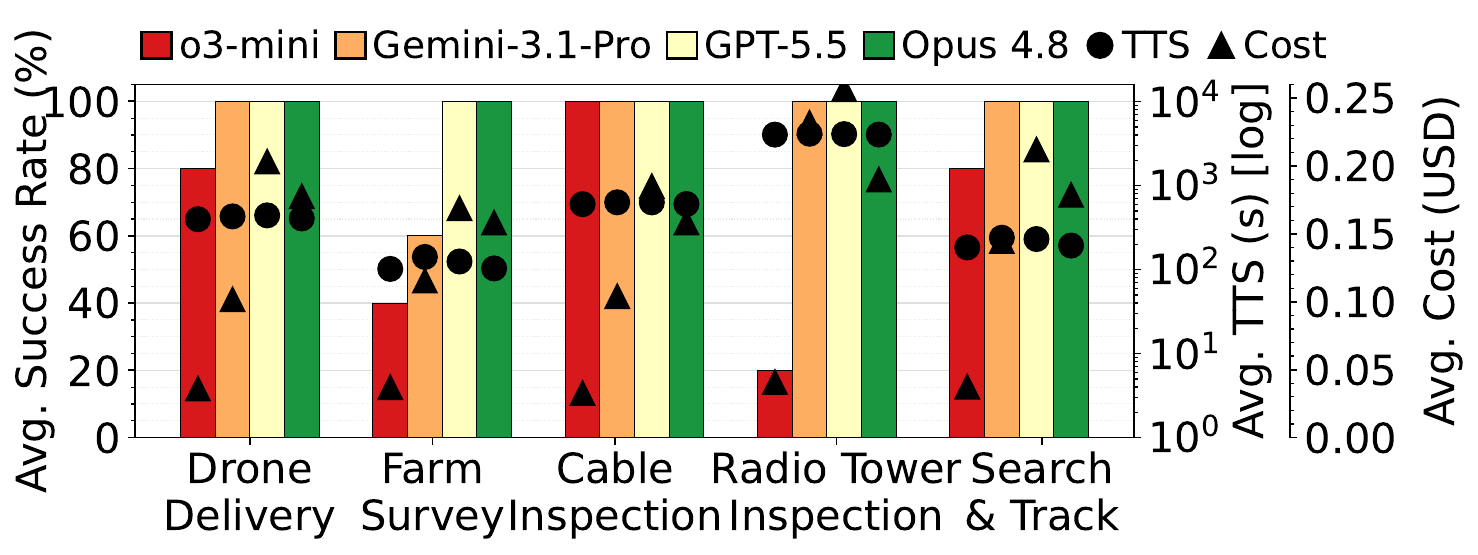}
\caption{Impact of LLM choice on analytical mission success, TTS, and generation cost.} 
\label{fig:multi-llm-performance} 
\end{minipage} 
\end{figure*}

\takeaway{LLM capability directly affects analytical mission reliability.} Fig.~\ref{fig:multi-llm-performance} compares \agen across LLMs with different reasoning capabilities. Frontier models, \textit{GPT-5.5} and \textit{Claude Opus 4.8}, achieve $100\%$ success with a corresponding $95\%$ Wilson CI of [0.57,1] across all evaluated analytical missions. \textit{Gemini-3.1-Pro}, a high-capability model, achieves a lower but still high average success rate of 92\% with a $95\%$ CI of [0.48,0.99], while \textit{OpenAI o3-mini}, a lower-cost compact reasoning model, is more variable, with mission-level success ranging from 20\% to 100\% and an average success rate of $64\%$ with a $95\%$ CI of [0.26,0.90]. The gap is most visible for Farm Survey and Radio Tower Inspection, where o3-mini succeeds in only $40\%$ and $20\%$ of runs with corresponding $95\%$ CIs of [0.12,0.77], and [0.04,0.62], respectively. Together with Fig.~\ref{fig:opus-sr}, these results show that guardrails are necessary even for frontier models.
The lower o3-mini success rate shows that model capability remains an independent factor in long-horizon planning, analytics selection, and spatial reasoning.

\takeaway{Lower-cost LLMs trade reliability for monetary cost.} The cost and TTS results in Fig.~\ref{fig:multi-llm-performance} show that o3-mini is substantially cheaper per attempt than frontier models, with a maximum generation cost of about \$\,$0.04$ per invocation compared to about \$\,$0.20$ for Opus 4.8 and \$\,$0.26$ for GPT-5.5. \textit{This makes repeated attempts economically viable for some missions}: a 20\% success rate implies an expected five attempts, or about \$\,$0.20$, while a 40\% success rate implies 2.5 attempts, or about \$\,$0.10$. However, repeated attempts increase end-to-end TTS because each failed program must still be generated and validated. Thus, lower-cost LLMs are useful when monetary cost is the primary constraint, while frontier models are preferable when single-shot reliability and wall-clock time are more important.

\para{Limitations and Reproducibility} \label{subsec:limitations} 
\agen improves single-shot generation for the evaluated, mostly sequential single-drone missions, but its evidence is scoped to the tested robots, worlds, analytics, backends and LLMs. We do not evaluate parallel or multi-drone composition, online replanning, or recovery from service and communication failures.
Failed baseline and ablated runs reveal recurring errors in waypoint semantics, platform assumptions, yaw or survey orientation, analytics selection, and world-rule compliance. 
Static, runtime, robot, and world guardrails reduce different subsets of these failures, but \agen is not a formal safety controller and does not guarantee robustness to dynamic obstacles, degraded sensors, stale world models, network failures, or adversarial prompts. Physical deployment should retain pre-flight checks, simulation validation, geofencing, emergency stops, and runtime safety monitors.


\section{Conclusions and Future Work}
\label{sec:conclusions}

In this paper, we presented \agen, an LLM-driven low-code framework for composing Drone-as-a-Service missions from natural-language definitions. \agen combines static API and constraint guardrails with dynamic robot, runtime and world information to generate executable mission programs. Across navigation and analytical missions, \agen succeeds in a single-shot for the evaluated missions with GPT-5.5 and Claude Opus 4.8, while ablations show higher analytical-mission success when dynamic context is included. The results also show practical tradeoffs among guardrail context, backend abstraction, LLM capability, TTS and monetary cost. 

Future work will strengthen \agen through closed-loop validation using simulation feedback and pre-deployment formal safety checks. We will evaluate stronger analytics-capable baselines, explore intermediate service representations for DaaS missions to improve composability and validation, and extend \agen to multi-drone missions and runtime adaptation under changing conditions.

\section*{Acknowledgment}
Generative AI tools were solely used to refine the clarity of author-written text in the paper. No new technical content, data or figures were generated. All such edits were carefully reviewed and verified by the authors.

\bibliographystyle{splncs04}
\bibliography{main}

\end{document}